\documentclass[accepted]{uai2026} 
                        
\usepackage[american]{babel}

\usepackage{natbib} 
\usepackage{mathtools} 
\usepackage{booktabs} 
\usepackage{tikz} 

\usepackage{multirow}
\usepackage{colortbl}
\usepackage[table,dvipsnames]{xcolor}
\usepackage{xspace} 
\newcommand{\Fern}{\texttt{Fern}\xspace}

\usepackage{colortbl}
\usepackage{xcolor}

\definecolor{BestTeal}{HTML}{D1E5F0}   
\definecolor{SecondTeal}{HTML}{EAF2F6} 
\definecolor{BadRust}{HTML}{FDDBC7}    

\newcommand{\first}[1]{\cellcolor{BestTeal}\textbf{#1}}
\newcommand{\second}[1]{\cellcolor{SecondTeal}#1}
\newcommand{\bad}[1]{\cellcolor{BadRust}\textcolor{black!60}{#1}}
\usepackage{amsfonts}
\usepackage{amssymb}

\title{Noise Titration: Exact Distributional Benchmarking for Probabilistic Time Series Forecasting}

\author[1]{\href{mailto:<qilin.wang@umontreal.ca>?Subject=Your UAI 2026 paper}{Qilin Wang}{}} 
\affil[1]{%
    Independent Researcher  
} 
  
\begin{document}
\maketitle

\begin{abstract}
  Modern time series forecasting is evaluated almost entirely through passive observation of single historical trajectories, rendering claims about a model's robustness to non-stationarity fundamentally unfalsifiable. We propose a paradigm shift toward interventionist, exact-statistical benchmarking. By systematically titrating calibrated Gaussian observation noise into known chaotic and stochastic dynamical systems, we transform forecasting from a black-box sequence matching game into an exact distributional inference task. Because the underlying data-generating process and noise variance are
mathematically explicit, evaluation can rely on exact
negative log-likelihoods and calibrated distributional tests
rather than heuristic approximations. To fully leverage this framework, we extend the Fern architecture into a probabilistic generative model that natively parameterizes the Symmetric Positive Definite (SPD) cone, outputting calibrated joint covariance structures without the computational bottleneck of generic Jacobian modeling. Under this rigorous evaluation, we find that state-of-the-art zero-shot foundation models behave consistently with the
context-parroting mechanism,
failing systematically under non-stationary regime shifts and
elevated noise. In contrast, Fern explicitly captures the invariant measure and multivariate geometry of the underlying dynamics, maintaining structural fidelity and statistically sharp calibration precisely where massive sequence-matching models collapse.
\end{abstract}

\section{Introduction}\label{sec:intro}

The Federal Reserve publishes the \textbf{dot plot} quarterly as part of its \textit{summary of economic projections}: the dot plot chart carries projection of interest rates for the current  and the next three calendar years, and for "longer run" (supposedly when the economy reaches a so-called \textit{stable state}). In real life, people pay little attention to those long-run forecasts. Instead, they outline different possible geopolitical scenarios and run forecasts under different scenarios. In other words, they actively \textit{model} the different non-stationary shocks and trust their forecaster to function normally in different scenarios.

Modern time series forecasting (TSF) is evaluated almost entirely in a passive, observational regime: models ingest a historical context and extrapolate a single realized future, while the underlying data-generating process (DGP) and the semantics of non-stationarity remain implicit. This makes robustness claims difficult to falsify: a ``hard'' dataset may reflect sensor artifacts, early-stopping quirks, or a single idiosyncratic realization rather than principled distribution shift. Consequently, we do not know whether state-of-the-art (SOTA) time series forecasters handle non-stationarity; we know only that they score well on benchmarks where non-stationarity happened to look a certain way. 

In experimental science, claiming a mechanism works requires specifying the intervention, controlling for confounds, and reporting calibrated statistical tests. Yet the dominant paradigm in TSF forecloses the very tools that make such rigor possible. \citet{wang2025friren} took a first step by arguing for \emph{intervention}: we use synthetic datasets whose data-generating process (DGP) is a known dynamical system, and we treat the type, timing, and magnitude of non-stationary shocks as \textit{explicit experimental variables}: a researcher then perturb either (1) parameters (2) state variables, or (3) switch regimes at controlled times. We observe if a method preserves predictive structure under counterfactuals (a range of varied paths) rather than chasing one single historical path. This interventionist view turns non-stationarity handling into a falsifiable, reproducible \textit{modelling} task of the forecaster.

As \cite{wang2025friren} puts it, this turns non-stationary shocks into an active modelling component, similar to making \emph{software tests for TSF models}: we outline potential edge cases, like we outline the geopolitical possibilities, and \textit{prove}, with full DGP and interventions knowledge, that the model functions properly under these cases. We \textit{actively discount} the value of ultra long single trajectory prediction, as we know shocks, chaos and stochastic noise eventually make such long shot pointwise forecast impossible; we also actively discount the value of what \cite{wang2025friren} termed \textit{historical emulation}, knowing well that different news, market obsessions etc could have resulted in a different trajectory; any model replicating the prediction either reach it by chance, or is overfit to that particular set of realized shocks.

\paragraph{A Statistical Approach to Prediction: The Noise Titration Protocol} 
Controlled DGPs make shock-handling claims \emph{falsifiable}; in this paper we propose a major expansion in this line of thinking: we make them \emph{inferrable}. By \textit{wrapping} calibrated Gaussian noise of known variance $\sigma^2$ around deterministic trajectories, which can be understood as some simulated post-hoc measurement noises, the target distribution is $\mathcal{N}(\mu_{\text{law}},\,\sigma^{2}I)$ by construction. 

Treating injected noise variance $\sigma_{\text{inj}}^2$ as an experimental dial, we gain two complementary, falsifiable perspectives on model capacity:
\begin{itemize}
    \item \textbf{Dialing Down (The Resolution Limit):} As the injected noise decreases ($\sigma \to 0$), we measure the model's \textit{epistemic precision}. By systematically unmasking the architecture's \textit{inherent} structural error, we quantify its \textit{exact lower bound}: if the forecaster's residuals pass a distributional
goodness-of-fit test at an injected noise level of $10^{-3}$ but
fail at $10^{-4}$, we explicitly establish the model's statistical
resolution limit at $10^{-3}$.
    \item \textbf{Dialing Up (The Robustness Threshold):} As we increase the noise level, we test the model's ability to extract the true invariant measure from severe \textit{aleatoric corruption}. This identifies the exact Signal-to-Noise Ratio (SNR) threshold where the model's manifold learning collapses into mean-guessing.
\end{itemize} 

We are \textit{not} forcing the object to fit a statistical hammer; in contrast, now the `wind-tunnel' interventionist experiments \textit{speaks the statistical inference language}. SOTA MSE competition on a single historical realization is replaced by \textbf{model performance in precise scientific terms}: under \textit{this} nonstationarity 
pattern, for \textit{this} dynamical system family, the model 
retains accuracy across $\sigma \in \{0.25, 1.0\}$ and degrades 
at $\sigma = 2.0$. A single evaluation loop \textit{describes} the non-stationarity we care, the noise structure (which \textit{can} be heteroskedastic by design) we care, and sigma resolution level we care. We give models a rigorous, statistically verified ``factory stamp'' of their robustness profile before they ever leave the laboratory. This connects prediction back to its statistical root as a tool for \textit{decision-making}, not an oracle that produces 
unqualified point forecasts, and forces the researcher to model the type of domain-specific non-stationarity for practical deployment.

The Fern model from \citep{wang2025friren} serves as the main model due to its \textit{unique properties}: it natively outputs a full Gaussian belief $\mathcal{N}(\mu_\theta,\Sigma_\theta)$ as well as the eigenvalues and eigenvectors of the covariance matrix. The parameterization makes closed-form $W_2$ distances, exact coverage tests and calibrated hypothesis testing much easier.
 
In this paper, we take three steps toward this paradigm:
\begin{enumerate}
    \item We extend Fern with a multivariate Gaussian NLL 
    objective, promoting its spectral parameterization from an 
    implicit geometric belief to a calibrated probabilistic 
    forecaster that natively outputs full covariance structure.
    \item We benchmark Fern against foundation models 
    (Chronos Bolt and Chronos 2~\citep{ansari2024chronos,ansari2025chronos2}, TimesFM 2.5~\citep{pmlr-v235-das24c-timefm}) across chaotic and nonstationary systems, 
    demonstrating order-of-magnitude improvements on chaotic 
    benchmarks.
    \item We conduct noise titration experiments that 
    characterize each model's resolution profile, providing a 
    falsifiable, quantitative alternative to single-realization 
    leaderboard ranking.
\end{enumerate}

\section{Method}
\paragraph{Quick intro to Fern}

We briefly recap the central idea of the Fern model proposed by
\citet{wang2025friren}: it views long-term time-series forecasting
as \emph{conditional manifold transport}: given a context window
$x_{1:L}$ (an evenly sampled time-delay embedding), predict a distribution over a future horizon $y_{1:H}$ by transporting probability mass along the system’s latent geometry. Real systems blur or disrupt this geometry through stochastic noise, chaotic forces, and regime-switching non-stationary shocks. 

Fern addresses this by directly modelling a \emph{instance-wise Brenier map} from a fixed Gaussian source $y_0\sim\mathcal N(0,I)$ to a Gaussian prediction $\mathcal N(\mu_\theta(x),\Sigma_\theta(x))$, conditional on information from $x$. By Brenier's theorem, the $W_2$-optimal map between Gaussians is affine with a symmetric positive (semi-)definite (SPD) Jacobian, so Fern  directly parameterizes the eigendecomposition that exists for every SPD matrix: For each horizon patch (and all patches are generated \textit{in parallel}), it draws $y_0\sim\mathcal N(0,I)$ and applies the affine map $y^* = U\Lambda U^\top (y_0+t_y)$;  $\Lambda_\theta\succeq 0$ is the diagonal matrix containing the eigenvalues and $U_\theta$ (parameterized by $R$ Householder reflections) are the rotational matrix whose columns are the eigenvectors.   

We briefly outline the mechanism and refer the reader to the original paper for details: via bidirectional coupling network, in the same way as in Augmented Normalizing Flow \cite{huang2020anf}, the context $x$ is encoded into a latent Gaussian $z$. A fresh drawn $z\sim \mathcal{N}(0,1)$ seeds the decoder, which generates per-patch spectral factors $(t,\Lambda,U)$ of an SPD transport. This dynamically generated, geometry-aware forecaster thus natively produces both mean ($\mu_\theta$) and covariance (via $\Sigma_\theta=U\Lambda U ^\top U\Lambda U ^\top=U\Lambda^2U^\top$) exploiting the SPD structure ($\Sigma=AA\top$ and $A=U\Lambda U\top$), while avoiding the $O(n^2)$ parameterization and $O(n^3)$ eigendecomposition costs of generic Jacobian modelling.

\paragraph{Gaussian Noise Injection and Exact NLL.}
We augment deterministic trajectories with observation noise $\varepsilon \sim \mathcal{N}(0, \sigma^2 I)$ of known variance. The target distribution becomes $\mathcal{N}(\mu_{\text{law}}, \sigma^2 I)$ by construction. Since Fern natively outputs $\mathcal{N}(\mu_\theta, \Sigma_\theta)$, both prediction and target live in the Gaussian family. This enables closed-form $W_2$ distances, exact coverage tests, and multivariate Negative Log-Likelihood (NLL) training without post-hoc variance estimation.

Under a standard point-prediction loss such as MSE or Huber, Fern's covariance $\Sigma_\theta = U\Lambda^2 U^\top$ remains an untrained byproduct of the architecture: it is a \textit{Gaussian belief} that stretch whatever amount that is necessary to minimize pointwise error. In this paper, we \textit{promote} it by using multivariate Gaussian NLL as the \textit{training objective}, decomposed in Fern's native eigenframe:
\begin{equation*}\label{eq:fern-nll}
  \mathcal{L}_{\mathrm{NLL}}
  = \frac{1}{2}\sum_{i=1}^{P}
    \Bigl[\,\log\lambda_i^2
    + \frac{r_i^2}{\lambda_i^2}\,\Bigr]
  + \mathrm{const},
  \qquad
  r = U^\top(y - \mu_\theta),
\end{equation*}
where $r_i$ is the residual projected onto the $i$-th eigendirection
and $\lambda_i$ is the corresponding eigenvalue. The $r_i^2/\lambda_i^2$ term penalizes overconfidence (eigenvalues
too small relative to the actual error), while the $\log\lambda_i^2$
term penalizes overdispersion (eigenvalues unnecessarily large). For fixed $r_i$, minimizing the per-direction term
$\log\lambda_i^2 + r_i^2/\lambda_i^2$ gives
$\frac{\partial}{\partial \lambda_i}\!\left(\log\lambda_i^2 + r_i^2/\lambda_i^2\right)
= \frac{2}{\lambda_i} - \frac{2r_i^2}{\lambda_i^3}=0
\;\Rightarrow\; \lambda_i^2 = r_i^2 \;\Rightarrow\; \lambda_i = |r_i|$ (since $\lambda_i\ge 0$).
Thus, the equilibrium $\lambda_i \approx |r_i|$ calibrates each
eigendirection to its empirical error magnitude. Therefore, we train
the \textit{full joint covariance structure} rather than marginal variances
alone. Because each forward pass draws a stochastic source
$y_0\sim\mathcal{N}(0,I)$, multiple samples ($S=8$ during
training) are used to reduce gradient variance when fitting
both $\mu_\theta$ and $\Sigma_\theta$ simultaneously.

Fern's transport structure admits natural extensions beyond the Gaussian family: its shift-rotate-scale operations preserve any elliptically contoured distribution. However, we restrict our focus to Gaussians in this work to enable exact, closed-form inference. Because Fern's spectral parameterization natively outputs the full covariance matrix $\Sigma_\theta$, exact calibration diagnostics, Wasserstein distances, and negative log-likelihoods are direct byproducts of the architecture rather than post-hoc approximations. Furthermore, because the architecture explicitly parameterizes an instance-wise optimal transport from the initial source noise to the prediction, it offers a strong geometric guarantee. 

We additionally compute whitened Mahalanobis distances as an omnibus calibration check; see Appendix~\ref{app:note:whitened} for details.

\section{Experiments and Results} 
\subsection{Setup} 
We benchmark \textsc{Chronos} (Chronos-Bolt, and Chronos-2) \citep{ansari2024chronos} and \textsc{TimesFM-2.5} \citep{pmlr-v235-das24c-timefm} as our primary baselines, as they represent the two dominant, state-of-the-art architectural paradigms in zero-shot time series foundation models: discrete LLM-based tokenization and continuous patched representations, respectively. \textsc{Fern} was previously shown to outperform pointwise
baselines on related benchmarks \citep{wang2025friren}. We forecast a horizon $H\in\{64,192\}$ with a fixed context window $L=336$ and report MSE for point-forecast accuracy, and CRPS for probabilistic accuracy. We follow \cite{wang2025friren} and provide component-wise SWD and EPT. Because large models are pre-trained and evaluated in a strictly zero-shot capacity, we report their performance using a single deterministic seed; for Fern we use 4 seeds $\{1955, 7, 20, 2023\}$. The Chronos models are built for best performances at length $\le 64$, the reader should keep that in mind for a fairer comparison.

\subsection{Performance Analysis}

\begin{table*}[t]
\centering
\footnotesize
\caption{\textbf{Forecasting MSE (H=64/192).} Models: fr (fern), c2 (chronos-2), tfm (timesfm-2.5-200m-pytorch), cb (chronos-bolt-base), cs (chronos-bolt-small). Lower is better. \first{Purple}: best; \second{Light Purple}: second; \bad{Orange}: diverged.}
\label{tab:synthetic-mse}
\setlength{\tabcolsep}{2.0pt}
\begin{tabular}{lrrrrrrrrrr}
\toprule
  & \multicolumn{2}{c}{fr} & \multicolumn{2}{c}{c2} & \multicolumn{2}{c}{tfm} & \multicolumn{2}{c}{cb} & \multicolumn{2}{c}{cs} \\
\cmidrule(r){2-3}
\cmidrule(lr){4-5}
\cmidrule(lr){6-7}
\cmidrule(lr){8-9}
\cmidrule(l){10-11}
Dataset & 64 & 192 & 64 & 192 & 64 & 192 & 64 & 192 & 64 & 192 \\
\midrule
\addlinespace[2pt]
\multicolumn{11}{l}{\emph{Standard Chaotic Dynamics}} \\
Rössler-Base
 & \first{0.024} & \first{0.016}
 & \second{1.04} & 12.7
 & 2.13 & \second{12.2}
 & 8.52 & \bad{108}
 & 7.67 & \bad{105} \\
Rössler-Param
 & \first{0.059} & \first{0.086}
 & \second{11.0} & \second{43.9}
 & 31.6 & 65.0
 & 65.0 & \bad{444}
 & 64.4 & \bad{607} \\
Lorenz-Base
 & \first{1.34} & \first{11.6}
 & \second{54.9} & 92.8
 & 61.4 & 92.4
 & 71.0 & 92.2
 & 71.3 & \second{91.1} \\
Lorenz-State
 & \first{1.82} & \first{10.6}
 & \second{53.7} & 89.0
 & 54.2 & \second{82.1}
 & 63.5 & 87.2
 & 64.7 & 87.0 \\
Lorenz-Param
 & \first{4.48} & \first{21.6}
 & \second{46.9} & 83.8
 & 48.3 & \second{78.6}
 & 61.0 & 86.4
 & 61.7 & 82.4 \\
Lorenz-Switch
 & \first{3.77} & \first{11.8}
 & \second{42.6} & \second{77.2}
 & 49.7 & 79.6
 & 58.6 & 83.3
 & 61.5 & 81.5 \\
Lorenz96-Base
 & \first{0.436} & \first{2.29}
 & \second{6.33} & \second{15.1}
 & 6.90 & 16.3
 & 9.43 & 18.4
 & 9.20 & 17.5 \\
Lorenz96-Switch
 & \first{1.58} & \first{7.06}
 & \second{9.27} & \second{18.8}
 & 10.1 & 20.7
 & 12.6 & 21.8
 & 12.5 & 21.1 \\
Chua-Base
 & \first{0.006} & \first{0.012}
 & \second{0.009} & \second{0.374}
 & 0.011 & 0.396
 & 0.033 & 0.642
 & 0.045 & 0.718 \\
Chua-Param
 & \first{0.008} & \first{0.008}
 & 0.019 & \second{0.682}
 & \second{0.018} & 0.784
 & 0.060 & 1.21
 & 0.079 & 1.41 \\
Chua-Switch
 & \first{0.004} & \first{0.019}
 & 0.011 & \second{0.421}
 & \second{0.010} & 0.489
 & 0.039 & 0.777
 & 0.049 & 0.826 \\
\addlinespace[2pt]
\multicolumn{11}{l}{\emph{Switching Linear Dynamical System}} \\
SLDS-Base
 & \first{2.55} & \first{2.89}
 & 3.35 & 3.72
 & \second{3.02} & \second{3.47}
 & 4.28 & 5.48
 & 3.51 & 4.09 \\
SLDS-Param
 & \first{2.18} & \first{2.23}
 & 2.62 & 2.68
 & \second{2.29} & \second{2.37}
 & 3.29 & 3.35
 & 3.13 & 3.27 \\
SLDS-Switch
 & \first{3.87} & \first{4.67}
 & 4.91 & 6.61
 & \second{4.43} & \second{5.61}
 & 6.27 & 8.57
 & 5.83 & 7.85 \\
\addlinespace[2pt]
\multicolumn{11}{l}{\emph{Seasonal AR Shocks}} \\
SAR-Base
 & 0.063 & \first{0.054}
 & \second{0.057} & \second{0.058}
 & \first{0.056} & 0.060
 & 0.059 & 0.066
 & 0.058 & 0.060 \\
SAR-Param
 & \first{0.342} & \first{0.326}
 & 0.373 & 0.399
 & \second{0.343} & \second{0.368}
 & 0.407 & 0.435
 & 0.382 & 0.399 \\
\addlinespace[2pt]
\multicolumn{11}{l}{\emph{Double-Well Potential}} \\
DW-Base
 & 0.056 & 0.063
 & 0.047 & 0.047
 & \first{0.047} & \first{0.047}
 & \second{0.047} & \second{0.047}
 & 0.047 & 0.047 \\
DW-Param
 & 0.624 & \first{0.769}
 & 0.630 & 0.981
 & \first{0.534} & \second{0.877}
 & 0.601 & 0.914
 & \second{0.599} & 0.934 \\
DW-Switch
 & \first{0.358} & \first{0.652}
 & 0.367 & 0.795
 & 0.379 & 0.882
 & \second{0.363} & \second{0.752}
 & 0.364 & 0.788 \\
\addlinespace[2pt]
\multicolumn{11}{l}{\emph{Ornstein--Uhlenbeck Diffusions}} \\
OU-Base
 & \first{0.195} & \first{0.203}
 & 0.219 & 0.238
 & \second{0.200} & \second{0.213}
 & 0.245 & 0.256
 & 0.228 & 0.247 \\
OU-Param
 & \second{0.217} & \first{0.204}
 & 0.219 & 0.238
 & \first{0.200} & \second{0.213}
 & 0.245 & 0.256
 & 0.228 & 0.247 \\
\bottomrule
\end{tabular}
\end{table*}

\begin{table*}[t]
\centering
\footnotesize
\caption{\textbf{Forecasting CRPS (H=64/192).} Models: fr (fern), c2 (chronos-2), tfm (timesfm-2.5-200m-pytorch), cb (chronos-bolt-base), cs (chronos-bolt-small). Lower is better. \first{Purple}: best; \second{Light Purple}: second; \bad{Orange}: diverged.}
\label{tab:synthetic-crps}
\setlength{\tabcolsep}{2.0pt}
\begin{tabular}{lrrrrrrrrrr}
\toprule
  & \multicolumn{2}{c}{fr} & \multicolumn{2}{c}{c2} & \multicolumn{2}{c}{tfm} & \multicolumn{2}{c}{cb} & \multicolumn{2}{c}{cs} \\
\cmidrule(r){2-3}
\cmidrule(lr){4-5}
\cmidrule(lr){6-7}
\cmidrule(lr){8-9}
\cmidrule(l){10-11}
Dataset & 64 & 192 & 64 & 192 & 64 & 192 & 64 & 192 & 64 & 192 \\
\midrule
\addlinespace[2pt]
\multicolumn{11}{l}{\emph{Standard Chaotic Dynamics}} \\
Rössler-Base
 & \first{0.090} & \first{0.071}
 & \second{0.206} & 1.09
 & 0.230 & \second{1.09}
 & 0.559 & 2.38
 & 0.585 & 2.69 \\
Rössler-Param
 & \first{0.095} & \first{0.144}
 & \second{0.394} & \second{1.67}
 & 0.562 & 1.81
 & 1.04 & 3.90
 & 1.08 & 4.87 \\
Lorenz-Base
 & \first{0.555} & \first{1.47}
 & \second{3.88} & 5.80
 & 4.31 & \second{5.79}
 & 4.82 & 5.97
 & 4.92 & 5.98 \\
Lorenz-State
 & \first{0.621} & \first{1.54}
 & \second{3.79} & 5.65
 & 4.04 & \second{5.43}
 & 4.56 & 5.84
 & 4.67 & 5.86 \\
Lorenz-Param
 & \first{0.707} & \first{2.23}
 & \second{3.30} & 5.33
 & 3.64 & \second{5.22}
 & 4.39 & 5.79
 & 4.50 & 5.69 \\
Lorenz-Switch
 & \first{0.908} & \first{1.57}
 & \second{3.00} & \second{5.08}
 & 3.54 & 5.17
 & 4.34 & 5.77
 & 4.55 & 5.73 \\
Lorenz96-Base
 & \first{0.295} & \first{0.823}
 & \second{1.28} & \second{2.28}
 & 1.31 & 2.40
 & 1.63 & 2.62
 & 1.63 & 2.59 \\
Lorenz96-Switch
 & \first{0.611} & \first{1.65}
 & \second{1.55} & \second{2.55}
 & 1.59 & 2.71
 & 1.87 & 2.85
 & 1.88 & 2.86 \\
Chua-Base
 & 0.046 & \first{0.053}
 & \second{0.040} & \second{0.265}
 & \first{0.034} & 0.269
 & 0.082 & 0.394
 & 0.100 & 0.426 \\
Chua-Param
 & \second{0.050} & \first{0.052}
 & 0.059 & \second{0.355}
 & \first{0.049} & 0.384
 & 0.113 & 0.550
 & 0.131 & 0.596 \\
Chua-Switch
 & \second{0.042} & \first{0.066}
 & 0.044 & \second{0.278}
 & \first{0.035} & 0.292
 & 0.090 & 0.430
 & 0.105 & 0.447 \\
\addlinespace[2pt]
\multicolumn{11}{l}{\emph{Switching Linear Dynamical System}} \\
SLDS-Base
 & \first{1.00} & \first{1.06}
 & 1.10 & 1.28
 & \second{1.03} & \second{1.15}
 & 1.23 & 1.42
 & 1.13 & 1.23 \\
SLDS-Param
 & \second{0.962} & \second{0.974}
 & 0.999 & 1.05
 & \first{0.932} & \first{0.951}
 & 1.11 & 1.14
 & 1.09 & 1.12 \\
SLDS-Switch
 & \first{1.20} & \first{1.31}
 & 1.33 & 1.71
 & \second{1.24} & \second{1.43}
 & 1.49 & 1.78
 & 1.43 & 1.69 \\
\addlinespace[2pt]
\multicolumn{11}{l}{\emph{Seasonal AR Shocks}} \\
SAR-Base
 & \first{0.144} & \first{0.134}
 & 0.148 & \second{0.150}
 & \second{0.146} & 0.151
 & 0.152 & 0.159
 & 0.150 & 0.153 \\
SAR-Param
 & \first{0.343} & \first{0.336}
 & 0.380 & 0.393
 & \second{0.363} & \second{0.377}
 & 0.404 & 0.414
 & 0.387 & 0.393 \\
\addlinespace[2pt]
\multicolumn{11}{l}{\emph{Double-Well Potential}} \\
DW-Base
 & 0.135 & 0.144
 & 0.133 & \second{0.132}
 & \first{0.132} & \first{0.132}
 & \second{0.132} & 0.135
 & 0.133 & 0.135 \\
DW-Param
 & 0.469 & \second{0.575}
 & 0.431 & 0.598
 & \first{0.403} & \first{0.564}
 & \second{0.422} & 0.577
 & 0.428 & 0.592 \\
DW-Switch
 & \second{0.322} & \first{0.504}
 & 0.323 & 0.514
 & 0.329 & 0.553
 & \first{0.320} & \second{0.513}
 & 0.326 & 0.525 \\
\addlinespace[2pt]
\multicolumn{11}{l}{\emph{Ornstein--Uhlenbeck Diffusions}} \\
OU-Base
 & \first{0.263} & \first{0.268}
 & 0.289 & 0.308
 & \second{0.276} & \second{0.286}
 & 0.306 & 0.315
 & 0.296 & 0.308 \\
OU-Param
 & \second{0.279} & \first{0.268}
 & 0.289 & 0.308
 & \first{0.276} & \second{0.286}
 & 0.306 & 0.315
 & 0.296 & 0.308 \\
\bottomrule
\end{tabular}
\end{table*}

Tables~\ref{tab:synthetic-mse} and \ref{tab:synthetic-crps} report MSE and CRPS across all experimental families at horizons H=64 and H=192. Several patterns emerge. \textbf{On chaotic systems, FERN outperforms all foundation models by one to two orders of magnitude}. On Lorenz-Base (H=64), FERN achieves MSE 1.34 against 54.9 (Chronos-2), 61.4 (TimeFM), and 71.0 (Chronos-Bolt-Base) --- a factor of 40–52×. The gap is similar on the easier task of Chua circuits, where FERN's MSE of 0.006 sits against 0.009–0.045. This pattern is consistent across CRPS: on Rössler-Base, FERN scores 0.090 versus 0.206–2.69 from baselines. Notably, many foundation model entries on chaotic benchmarks are marked diverged (orange), indicating that the model's output drifted to uninformative levels — a failure mode FERN never exhibits.

The advantage persists and often \textit{widens under nonstationarity}. Across parameter drift (-Param), state-dependent (-State), and regime switching (-Switch) variants of each chaotic system, FERN maintains low error while foundation models degrade further. On Lorenz-Param (H=192), FERN scores MSE 21.6 versus 83.8 (Chronos-2) and 78.6 (TimeFM). This is consistent with the context parroting hypothesis of \cite{zhang2025contextparroting}: foundation models forecast by matching motifs in the context window, but when the underlying attractor changes mid-sequence, past motifs lose their predictive validity. FERN, with \textit{instance-wise} transport generation, does not rely on this mechanism but rather analyzes the context geometry. Among foundation models, Chronos-2 is generally the strongest performer, followed by TimeFM. This aligns with prior findings that Chronos's cross-entropy training and tokenization scheme implicitly encourage parroting, the most effective zero-shot strategy on recurrent attractors. TimeFM and Chronos-Bolt, trained with MSE loss, tend to regress toward the mean at longer horizons --- visible in their sharply inflated H=192 scores on chaotic benchmarks.

On non-chaotic families, the gap \textit{narrows considerably}. For Double-Well Base (H=64), TimeFM achieves 0.047 versus FERN's 0.056; on SAR-Base, FERN's CRPS of 0.144 edges TimeFM's 0.146. Ornstein–Uhlenbeck processes show Fern leading but by a smaller margin. This is expected: stationary systems have smoother, more periodic, or mean-reverting structures where the recent past is highly representative of the near future, making them genuinely easier to solve via context parroting \citep{zhang2025contextparroting}. However, chaotic and non-stationary systems are fatally deceptive: sensitive dependence and sudden parameter drifts mean that historical similarity deteriorates very quickly. This suggests the competitive advantage of a trained spectral model is its ability to track genuine geometry—maintaining structural fidelity precisely in the regimes where context parroting breaks down.

Horizon scaling further separates the approaches. FERN's MSE ratio from H=64 to H=192 on Lorenz-Base is 8.6×, while Chronos-2 grows 1.7× and TimeFM 1.5× — but from a much higher baseline. The foundation models' flatter scaling reflects their tendency to collapse to the marginal mean at long horizons rather than tracking the true trajectory, which artificially limits error growth at the cost of informativeness. FERN's error growth reflects genuine sensitivity to chaotic divergence while maintaining trajectory tracking far longer.

\subsection{The Noise Titration Protocol}

As discussed in the Methods section, to evaluate Fern's distributional calibration, we systematically titrated observation noise into the Rössler and Double Well systems. To recap, for each test window $t$, Fern outputs a predictive mean 
$\mu_\theta(x_t)$ and an eigendecomposition 
$\Sigma_\theta(x_t) = U \Lambda^2 U^\top$. We project the 
innovation $\varepsilon_t = y_t - \mu_\theta(x_t)$ into the 
eigenbasis:
\[
  \tilde{\varepsilon}_{t,i} = (U^\top \varepsilon_t)_i, \qquad
  z_{t,i} = \frac{\tilde{\varepsilon}_{t,i}}{\lambda_i}.
\]
This yields 
three levels of diagnostics.

\paragraph{Probability Integral Transform (PIT) Plots} First, we use probability integral transform (PIT) evolution plot to show the intuition of titration. If a model predicts a CDF $F_\theta$ and the true value is $y$, then $u = F_\theta(y)$ should be distributed as $\mathsf{Unif}(0,1)$ if the model is calibrated. It's the fundamental theorem of calibration --- any correctly specified probabilistic model for a continuous $y$ produces uniform PIT values. A histogram of u values tells you instantly: flat = calibrated, U-shaped = underdispersed (too narrow), hump-shaped = overdispersed (too wide).

\begin{figure*}[ht]
    \centering
    \includegraphics[width=\textwidth]{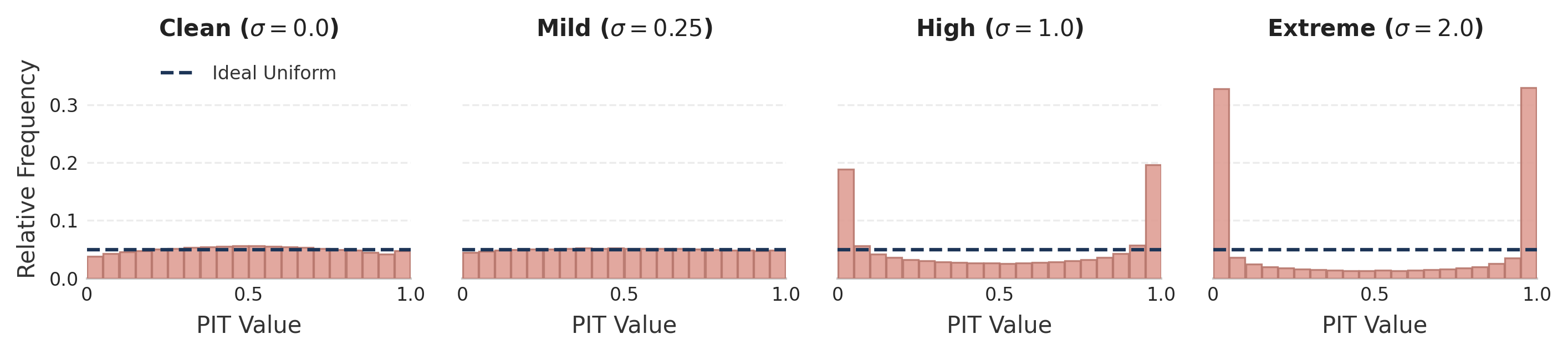}
    \caption{\textbf{Probability Integral Transform (PIT) Evolution under Noise Titration.} PIT histograms for Fern on the Double Well system. A perfectly calibrated model exhibits a uniform distribution (red dashed line). At clean and mild noise levels, the predictive distribution is well-calibrated. At extreme noise ($\sigma=2.0$), the histogram forms a severe U-shape, visually confirming that the model becomes pathologically under-dispersed (overconfident) when the aleatoric noise scale overwhelms the manifold dynamics.}
    \label{fig:pit_titration}
\end{figure*}

 The PIT plot as \ref{fig:pit_titration} shows, Double Well system at zero (but still with built-in system noise) and $\sigma=0.25$ post-hoc noise are very well calibrated; at $sigma=1.0$ it starts to collapse: when the true y falls in the tails (which happens more often than the model expects), $F_\theta(y)$ gets pushed toward 0 or 1, meaning the variances of the model (note that we place a $[0,5.5]$ limit on the scales of Fern) could be too small and the model ends up being too confident. At $\sigma=2$ the model is probably unsuited for the task. So we have a \textit{precise} description of \textit{robustness profile} against a meta-stable system under different levels of iid Gaussian measurement noise (again, non-iid noise structure \textit{can} be specified).


\paragraph{Shapiro-Wilk, FDR correction, and Coverages}
We move on to showcase another suite of statistical tools. Under our titration protocol the injected Gaussians are iid, so single dimension should not reject the null hypothesis under Shapiro-Wilk (SW) that $z_{t,i} \sim \mathcal{N}(0,1)$ i.e. the residual is normal, when the model is well-calibrated. Crucially, the injected noise level $\sigma$ is \textit{never passed explicitly} to the goodness-of-fit tests; it enters \textit{implicitly} through the data and must be natively absorbed by Fern's learned geometry. Equivalently, $m_t = \sum_i z_{t,i}^2 \sim \chi^2_d$ under $H_0$, connecting the per-direction tests to the classical Mahalanobis diagnostic. The Mahalanobis distance collapses into a scalar that is primarily sensitive to overall
location and scale, whereas per-direction Shapiro-Wilk tests \emph{shape} and invariant to location and scale. Using both
lets us disentangle shape errors from mis-centering and variance
miscalibration. Yet, with hundreds of dims, the simultaneous hypothesis testing requires adjustment, and we use the \textbf{false discovery rate} $q=0.05$ via Benjamini-Hochberg (BH-FDR) ~\cite{benjamini1995fdr} adjustments, as the Bonferroni~\cite{hochberg1987multiple} procedures are too conservative and the single-minded focus on Type I error is counterproductive: the null hypothesis is good model calibration, meaning a Type I error (reject null when it is true) is a false alarm that falsely penalizes a valid model, whereas Type II errors wrongly accept a broken, uncalibrated model as `good' when it actually collapses. By using BH-FDR, we ensure that among all spatial dimensions flagged as uncalibrated, at most $5\%$ are false alarms triggered by finite-sample noise rather than genuine manifold collapse. 

Such exact multivariate calibration diagnostics appears rarely in TSF previously, mainly due to architectural limitations: mean, covariance, eigenvectors and eigenvalues are generally too expensive to compute. Yet, relying solely on BH–FDR corrected normality tests on residuals is limited: as we inject higher noise levels, the noise dominates the dynamics, so the residuals become more and more trivially Gaussian regardless of whether Fern learned anything useful. The test passes for the wrong reason. We combine them with \textbf{empirical coverages}: fraction of test points where the true $y_t$ falls inside the model's predicted 50\% central interval. Concretely, the model predicts $\mathcal{N}(\mu_\theta, \Sigma_\theta)$,  we check if $y_t$ falls within the  $[\hat{q}_{0.25}, \hat{q}_{0.75}]$ quantiles of that distribution.

Table~\ref{tab:calibration} reports these diagnostics: empirical coverage for nominal 50\% and 90\% prediction intervals and Shapiro-Wilk (SW) pass rate of normality test with FDR adjustments. At mild noise ($\sigma=0.25$), the model maintains near-perfect coverage. Under extreme aleatoric corruption ($\sigma=2.0$), the noise dominates the dynamics (SW pass rate approaches $1.0$), but the model's learning collapses, causing severe under-coverage of the true variance. This is \textit{not} a contradiction, but a feature: while Shapiro–Wilk focuses purely on Gaussian shape (being invariant to location and scale), empirical coverage is dominated by location and variance calibration; using them together lets us disentangle shape errors from scale/centering errors and obtain a more complete calibration diagnosis. The joint signature (high pass rate, collapsed coverage) is only diagnosable because $\sigma$ is known by construction --- on observational data, a practitioner seeing high SW pass rates could have incorrectly conclude the model is well-calibrated.

\begin{table*}[t]
\centering
\footnotesize 
\setlength{\tabcolsep}{4pt} 
\caption{\textbf{Coverages and Normality Tests}
Empirical coverage at nominal 50\% and 90\% levels, and Shapiro--Wilk (SW) 
shape pass rate under BH FDR ($q{=}0.05$) across time 
dimensions, for Fern on Double Well (SDE) and R\"ossler (ODE) systems. 
\colorbox{gray!20}{Shaded} cells: $\sigma{=}0$ distributional coverage is \textit{undefined} as ODE targets are deterministic 
point masses. Bold: near-nominal 
calibration at $\sigma{=}0.25$.}
\label{tab:calibration}
\begin{tabular}{ll cccc ccc}
\toprule
& & \multicolumn{4}{c}{\textbf{Double Well (Stochastic SDE)}} & \multicolumn{3}{c}{\textbf{Rössler (Deterministic ODE)}} \\
\cmidrule(lr){3-6} \cmidrule(lr){7-9}
\textbf{Metric} & $\boldsymbol{\sigma}$ & Base & Param Shock & Switch & \textit{Ideal} & Base & Param Shock & \textit{Ideal} \\
\midrule
\multirow{4}{*}{\shortstack[l]{\textbf{Empirical}\\\textbf{Coverage}\\\textbf{(Nominal 50\%)}}} 
  & 0.00 & 0.535 & 0.481 & 0.476 & 0.50 & \cellcolor{gray!20}0.945 & \cellcolor{gray!20}0.920 & 0.50 \\
  & 0.25 & \textbf{0.521} & \textbf{0.491} & \textbf{0.480} & 0.50 & \textbf{0.635} & \textbf{0.605} & 0.50 \\
  & 1.00 & 0.271 & 0.248 & 0.247 & 0.50 & 0.263 & 0.259 & 0.50 \\
  & 2.00 & 0.143 & 0.137 & 0.138 & 0.50 & 0.138 & 0.138 & 0.50 \\
\midrule
\multirow{4}{*}{\shortstack[l]{\textbf{Empirical}\\\textbf{Coverage}\\\textbf{(Nominal 90\%)}}} 
  & 0.00 & 0.908 & 0.856 & 0.860 & 0.90 & \cellcolor{gray!20}0.998 & \cellcolor{gray!20}0.991 & 0.90 \\
  & 0.25 & \textbf{0.906} & \textbf{0.871} & \textbf{0.865} & 0.90 & \textbf{0.930} & \textbf{0.910} & 0.90 \\
  & 1.00 & 0.600 & 0.556 & 0.554 & 0.90 & 0.585 & 0.577 & 0.90 \\
  & 2.00 & 0.338 & 0.327 & 0.326 & 0.90 & 0.330 & 0.328 & 0.90 \\
\midrule
\multirow{4}{*}{\shortstack[l]{\textbf{SW Shape}\\\textbf{Pass Rate}\\\textbf{(FDR $q{=}0.05$)}}} 
  & 0.00 & 0.323 & 0.448 & 0.497 & 1.00 & 0.024 & 0.000 & 1.00 \\
  & 0.25 & 0.660 & 0.550 & 0.613 & 1.00 & 0.364 & 0.130 & 1.00 \\
  & 1.00 & 0.788 & 0.905 & 0.828 & 1.00 & 0.529 & 0.454 & 1.00 \\
  & 2.00 & 0.943 & 0.988 & 0.984 & 1.00 & 0.659 & 0.565 & 1.00 \\
\bottomrule
\end{tabular}
\end{table*}


\paragraph{Foundation Models under Noise Titration}
\label{sec:titration-comparison}

The noise titration protocol does not only characterize Fern---it stress-tests \textit{any} model. Table~\ref{tab:titration} reports CRPS at H=192 for Fern, Chronos-2, and Chronos-Bolt-Base across four noise levels on Rössler and Double Well, averaged over trajectory realizations.

\begin{table}[t]
\centering
\footnotesize
\setlength{\tabcolsep}{4pt}
\caption{\textbf{CRPS under noise titration.} \Fern\ at H\,=\,192;
foundation models at H\,=\,64 and H\,=\,192.
Systems: R\"{o}ssler-Base (RoB), Double-Well-Base (DWB),
Double-Well-Param (DWP). R\"{o}ssler-Param follows the same pattern as RoB.
\textbf{Bold}: best per row.}
\label{tab:titration}
\begin{tabular}{llccccc}
\toprule
 & & \multicolumn{1}{c}{\textbf{FR}} & \multicolumn{2}{c}{\textbf{C2}} & \multicolumn{2}{c}{\textbf{CB}} \\
\cmidrule(lr){3-3}\cmidrule(lr){4-5}\cmidrule(lr){6-7}
\textbf{Sys} & $\boldsymbol{\sigma}$ & 192 & 64 & 192 & 64 & 192 \\
\midrule
\multicolumn{7}{l}{\emph{Chaotic}}\\
RoB & 0.00 & \textbf{0.055} & 0.206 & 1.089 & 0.559 & 2.377 \\
 & 0.25 & \textbf{0.350} & 0.409 & 1.314 & 0.634 & 2.096 \\
 & 1.00 & \textbf{0.838} & 0.939 & 1.814 & 1.122 & 2.584 \\
 & 2.00 & \textbf{1.584} & 1.584 & 2.374 & 1.717 & 3.111 \\
\addlinespace[2pt]
\multicolumn{7}{l}{\emph{Stochastic}}\\
DWB & 0.00 & 0.220 & 0.133 & \textbf{0.132} & 0.132 & 0.135 \\
 & 0.25 & 0.270 & \textbf{0.205} & 0.205 & 0.205 & 0.209 \\
 & 1.00 & 0.683 & \textbf{0.637} & 0.641 & 0.637 & 0.650 \\
 & 2.00 & 1.409 & 1.254 & 1.261 & \textbf{1.254} & 1.280 \\
\addlinespace[2pt]
DWP & 0.00 & \textbf{0.512} & 0.431 & 0.598 & 0.422 & 0.577 \\
 & 0.25 & \textbf{0.496} & 0.473 & 0.613 & 0.448 & 0.592 \\
 & 1.00 & \textbf{0.789} & 0.771 & 0.848 & 0.764 & 0.853 \\
 & 2.00 & 1.470 & 1.339 & 1.398 & \textbf{1.327} & 1.391 \\
\bottomrule
\end{tabular}
\end{table}

Note the titration is run on different settings than benchmarking and ablation: see App.\ref{app:note:titration} for details. This is meant for discovery, refer to the main tables for controlled benchmark. 

Two patterns emerge. On Rössler (RoB), both foundation models degrade monotonically, suggesting the context parroting break down in Chaotic regimes even if extraordinary amount of \textit{blurring of geometry}: even at $\sigma=2.0$, where aleatoric noise dominates, Fern's CRPS of 1.58 remains on par with Chronos-2 at len-$64$. Stationarity matters: Chronos leads with \textit{base} Double Well (DWB), but not (at len-$192$) when nonstationarity kicks in (DWP). Yet, The \textit{most striking empirical finding} emerges from the MSE column for Chronos-Bolt on Rössler (H=192): MSE \emph{decreases} as noise increases, from 107.7 at $\sigma=0$ to 44.3 at $\sigma=0.25$ and 24.7 at $\sigma=1.0$. Adding observation noise to a chaotic trajectory \emph{improves} Chronos-Bolt's point accuracy. This is not a calibration artifact---it is a direct signature of context parroting. The model's blurry motif-matched output is a poor fit to the precise clean attractor, but when additive Gaussian noise smears the target, the distance between parroted output and noisy truth accidentally shrinks. A static benchmark at $\sigma=0$ would record MSE=107.7 and conclude the model fails on Rössler; it would never reveal that the model was inadvertently benefiting from noise artifacts in other evaluation conditions. The titration protocol exposes this in a single sweep.

\paragraph{Conformal Prediction under Non-Stationarity}

The titration approach provides a critical advantage over other uncertainty quantification standards. Modern probabilistic forecasting relies heavily on \textbf{Conformal Prediction (CP)} to generate valid prediction intervals. However, CP fundamentally requires the assumption of \textbf{exchangeability} between the training, validation, and test sets. Under the severe non-stationary shocks---such as regime switches and parameter drift, exchangeability is outright violated (even with remedies such as EnbPI \citep{xu2021conformal}, which provide distribution-free coverage via block bootstrap). By wrapping \textit{known} dynamical systems in Gaussian noise of \textit{known} variance, our protocol \textit{bypass} exchangeability: the Gaussian DGP is an engineered truth, not a convenience assumption. This enables exceptionally sharp, exact statistical bounds.

As a concrete stress test for conformal methods, we compare Fern's
Gaussian intervals to EnbPI \citep{xu2021conformal} on the
OU-ParamShock system (Table~\ref{tab:enbpi_ou_param}). Here the
parameter drift explicitly violates exchangeability. Even with a
block bootstrap tailored to time series (block length 48, bootstrap samples $B{=}5$),
EnbPI's nominal 90\% intervals collapse to 67.5\% empirical coverage (95\% interval $[0.668, 0.683]$). Fern's
fern\_nll-trained Gaussian belief on the same experiment attains
85.3\% coverage (95\% CI $[0.837, 0.870]$), i.e., mild
under-dispersion but nowhere near the 20+ point failure of EnbPI.
This illustrates that, once exchangeability is broken by regime
shifts, conformal guarantees no longer hold even with block
bootstrapping, whereas in our controlled DGP setting we can still
evaluate and improve calibration directly via likelihood and
coverage.

\begin{table}[ht]
\centering
\footnotesize
\setlength{\tabcolsep}{4pt}
\begin{tabular}{lcc}
\toprule
\textbf{Metric} & \textbf{Fern} & \textbf{EnbPI} \\
\midrule
Coverage (90\%) & 0.853 & 0.675 \\
95\% CI & [0.837, 0.870] & [0.668, 0.683] \\
$|\,\widehat{\mathrm{cov}} - 0.90\,|$ & \textbf{0.047} & 0.225 \\
\bottomrule
\end{tabular}
\caption{\textbf{Conformal EnbPI vs. Generative Fern Calibration on OU-ParamShock.} 
Comparison of nominal 90\% coverage under a parameter drift that violates exchangeability. Fern is evaluated via its native NLL-trained Gaussian belief ($\sigma=0.25$). EnbPI wraps the same base model using block-wise resampling ($L_{\text{block}}=48$, $B=5$). Despite its distribution-free guarantees under exchangeability, EnbPI severely under-covers on this non-stationary process, whereas Fern dynamically adapts.}
\label{tab:enbpi_ou_param}
\end{table}

\subsection{Ablation}

We tried three different training objectives. We leave the details of using diagonal NLL (the naive NLL that doesn't presume a full covariance structure) to Appendix~\ref{app:tab:ablation-mse} and App.~\ref{app:tab:ablation-crps}. Table~\ref{tab:ablation-loss} isolates the effect of the training objective: CRPS training (with training sample = 8, evaluation sample = 96) produces sharper predictions but leaves the eigenframe uncalibrated; Fern-NLL training (with training sample = 4, evaluation sample = 32) explicitly targets eigenvalue calibration, enabling the distributional diagnostics above at the cost of increased predictive spread.

\begin{table}[t]
\centering
\scriptsize 
\setlength{\tabcolsep}{4pt}
\caption{\textbf{Ablation: CRPS vs.\ NLL training objective.}
Both variants use Fern's spectral architecture with identical seeds $\{7, 1955\}$.
\textbf{Bold}: better per cell.}
\label{tab:ablation-loss}
\begin{tabular}{lrrrr}
\toprule
 & \multicolumn{2}{c}{\textbf{FERN-CRPS}} & \multicolumn{2}{c}{\textbf{FERN-NLL}} \\
\cmidrule(lr){2-3}\cmidrule(lr){4-5}
Dataset (H) & MSE & CRPS & MSE & CRPS \\
\midrule
\addlinespace[2pt]
\multicolumn{5}{l}{\emph{Chaotic Systems}} \\
Lorenz-Base (64)   & \textbf{0.830} & \textbf{0.371} & 1.341 & 0.555 \\
Lorenz-Base (192)  & \textbf{5.920} & \textbf{0.783} & 11.61 & 1.471 \\
Lorenz-Param (64)  & \textbf{4.305} & \textbf{0.613} & 4.480 & 0.707 \\
Lorenz-Param (192) & \textbf{18.14} & \textbf{1.484} & 21.58 & 2.226 \\
Lorenz-Switch (64) & \textbf{2.275} & \textbf{0.380} & 3.772 & 0.908 \\
Lorenz-Switch(192) & \textbf{6.622} & \textbf{0.749} & 11.77 & 1.571 \\
Rössler-Base (64)  & \textbf{0.0087} & \textbf{0.0477} & 0.0243 & 0.0904 \\
Rössler-Base (192) & 0.0345 & 0.0775 & \textbf{0.0159} & \textbf{0.0712} \\
Chua-Base (64)     & \textbf{0.0007} & \textbf{0.0240} & 0.0056 & 0.0462 \\
Chua-Base (192)    & \textbf{0.0108} & \textbf{0.0443} & 0.0120 & 0.0532 \\
\addlinespace[2pt]
\multicolumn{5}{l}{\emph{Stochastic / Smooth Systems}} \\
DW-Base (64)       & \textbf{0.0485} & \textbf{0.1245} & 0.0562 & 0.1353 \\
DW-Base (192)      & \textbf{0.0478} & \textbf{0.1231} & 0.0628 & 0.1439 \\
DW-Param (64)      & \textbf{0.5821} & \textbf{0.4365} & 0.6239 & 0.4687 \\
DW-Param (192)     & \textbf{0.7465} & \textbf{0.4857} & 0.7688 & 0.5748 \\
SAR-Base (64)      & \textbf{0.0534} & \textbf{0.1326} & 0.0633 & 0.1444 \\
SAR-Base (192)     & \textbf{0.0531} & \textbf{0.1318} & 0.0544 & 0.1337 \\
\bottomrule
\end{tabular}
\end{table}

\section{Related Work and Discussion}

\paragraph{Forecasting Architectures}
Long-term time series forecasting has seen rapid architectural evolution, from Transformer-based models \citep{vaswani2017attention, zhou2021informer, wu2021autoformer} to linear methods \citep{zeng2023transformersAAAI, nie2023patchtst} and frequency-domain approaches \citep{yi2023frequencydomain}. Foundational critiques \citep{zeng2023transformersAAAI, bergmeir2023foresight} have repeatedly demonstrated that simple linear models and even naive last-value predictors can match or outperform complex architectures, suggesting that patching and smoothing---not architectural sophistication---drive most empirical gains on standard benchmarks. A recent parallel trend pre-trains large models on diverse corpora for zero-shot forecasting. Chronos \citep{ansari2024chronos} tokenizes time series and trains via cross-entropy, while TimesFM \citep{pmlr-v235-das24c-timefm} and Moirai \citep{woo2024unified} use patched Transformer architectures with MSE or mixture-distribution objectives. \citet{zhang2025contextparroting} showed that these models forecast chaotic systems primarily through \emph{context parroting}: copying motif continuations from the context window. While effective on stationary attractors, this mechanism breaks severely under the regime shifts and parameter drift that define our non-stationary benchmarks.

\paragraph{Evaluation Methodology.}
Evaluation remains narrow: models are typically ranked by MSE on fixed historical splits, making claims about non-stationarity handling difficult to falsify. Alongside Fern~\citep{wang2025friren}, several recent works have raised parallel concerns and proposed suggestions on the three themes: domain-specific modelling, synthetic benchmark, and stress testing. \citet{ma2026position} argues that the ``one-architecture-fits-all'' benchmark culture is reaching diminishing returns and obscures domain-specific requirements, calling for evaluation and modelling that are aligned with the semantics of the target domain. In the same spirit, \citet{wang2025aries} introduces a synthetic pattern suite and a data-property-driven analysis/recommendation pipeline, emphasizing that model rankings depend strongly on which temporal relations (e.g., periodicity, coupling, non-stationarity) are present by construction. \citet{tan2025syntsbench} similarly advocates programmable synthetic benchmarks that isolate temporal pattern learning and stress-test robustness to irregularities (noise/anomalies), complementing pure leaderboard evaluation. Our noise titration protocol is consistent with these directions but targets a distinct gap: by injecting observation noise of known variance into controlled dynamical systems, we turn robustness under shocks into an explicit experimental dial and enable likelihood-based, distributional calibration checks beyond point-error metrics. 

\paragraph{Discussion and Future Directions: Beyond Gaussians.}
While we restrict our analysis to Gaussian noise to enable exact, closed-form NLL and $W_2$ inference, Fern's transport structure admits natural extensions. Because the transport maps are an affine composition of ``shift, rotate, scale, and rotate-back,'' \textit{any} distribution closed under affine transformations remains distributionally invariant. This includes the entire elliptical family of distributions, such as multivariate Student-t, Cauchy, and Laplace distributions, as well as Lévy symmetric alpha-stable distributions. If we omit the orthogonal rotations and limit the Brenier map strictly to scale and shift (a simple affine coupling layer), independent versions of the Laplace, Logistic, and Student-t distributions are also supported. Consequently, modeling severe heavy-tailed distributions—a fundamental challenge in financial forecasting—can be explicitly handled under this paradigm in future work simply by substituting the base noise profile.

\clearpage
\appendix

\section*{App: Note on Titration Comparison}
\label{app:note:titration}

For Fern, Rössler is run on one trajectories (fixed) plus three noise realization for each model seed; for Double Well it is three trajectories realization plus single noise realization for each model seed. For Chronos models, we run one trajectory realization plus one noise realization, and built for best performances at length $\le 64$.

\section{Mahalanobis Diagnostic}
\label{app:note:whitened}
\paragraph{Whitened Innovations and the Mahalanobis \texorpdfstring{$\chi^2$}{chi-square} Diagnostic}

Under our exact-inference benchmarking paradigm, the target data-generating process is explicitly defined as $y_t = F(x_t) + \varepsilon_t^\star$, where $\varepsilon_t^\star \sim \mathcal{N}(0, \Sigma^\star(x_t))$. Given a context $x_t$, the ideal forecaster $H_0$ must perfectly recover both the true drift and the underlying geometry: $\mu_\theta(x_t) = F(x_t)$ and $\Sigma_\theta(x_t) = \Sigma^\star(x_t)$.

To rigorously evaluate whether a model has achieved $H_0$, we analyze the prediction innovations, defined as $\varepsilon_t := y_t - \mu_\theta(x_t)$. If the predicted covariance is perfectly calibrated, the \emph{whitened} innovations must follow an isotropic standard Gaussian distribution.  

For generic Gaussian models, this whitening requires computing an expensive Cholesky decomposition or matrix inversion of $\Sigma_\theta$. However, because Fern natively parameterizes the eigendecomposition $\Sigma_\theta(x_t) = U_t \Lambda_t^2 U_t^\top$, we can compute the whitened residual $r_t$ trivially by projecting the innovation into the predicted eigenbasis and scaling by the inverse eigenvalues:
\begin{equation}
    r_t \;:=\; \Lambda_t^{-1} U_t^\top \varepsilon_t
\end{equation}
Under the null hypothesis $H_0$, the projected residuals must be strictly standard normal: $r_t \stackrel{\text{i.i.d.}}{\sim} \mathcal{N}(0, I_d)$. Equivalently, the squared Mahalanobis distance, computed efficiently as the squared $\ell_2$-norm of the projected residual, must follow a Chi-squared distribution with $d$ degrees of freedom:
\begin{equation}
    m_t \;:=\; \|r_t\|_2^2 \;=\; \sum_{i=1}^d \left(\frac{\tilde{\varepsilon}_{t,i}}{\lambda_{t,i}}\right)^2 \;\stackrel{\text{i.i.d.}}{\sim}\; \chi^2_d
\end{equation}
where $\tilde{\varepsilon}_t = U_t^\top \varepsilon_t$. To prevent numerical instability in nearly degenerate ellipsoids, the predicted eigenvalues $\lambda_{t,i}$ are clamped to a strict positive lower bound $\lambda_{\min} > 0$ during computation.

This formulation provides a powerful, scalar omnibus goodness-of-fit diagnostic for the model's entire conditional joint distribution.  For each experiment, we compute the empirical distribution of $\{m_t\}_{t=1}^T$ on the held-out test set and evaluate it against the theoretical $\chi^2_d$ law using a one-sample Kolmogorov-Smirnov (K-S) test. A rejection of the null hypothesis explicitly flags a mis-specification in the model's predictive mean, covariance geometry, or distributional shape, yielding a statistically sharp boundary for model failure.

\section*{App: Full Ablation Tables}

\begin{table*}[t]
\centering
\footnotesize
\caption{\textbf{Ablation Study: Forecasting MSE (H=64/192).} Comparison of Fern trained with different objective functions: CRPS loss (\texttt{CRPS}) with train sample = 8, evaluation samples = 96, the \textit{diagonal} Negative Log-Likelihood implementation (\texttt{NLL-old}) with train sample = 8, evaluation samples = 96, and Fern Gaussian NLL with full covariance matrix (\texttt{NLL-new}), with train sample = 4, evaluation samples = 32. Missing entries (-) denote datasets omitted in the partial CRPS run. Lower is better. Bold indicates the best result among the Fern variants.}
\label{app:tab:ablation-mse}
\setlength{\tabcolsep}{4.0pt}
\begin{tabular}{l ccc ccc}
\toprule
 & \multicolumn{3}{c}{\textbf{H=64}} & \multicolumn{3}{c}{\textbf{H=192}} \\
\cmidrule(lr){2-4} \cmidrule(lr){5-7}
\textbf{Dataset} & \textbf{CRPS} & \textbf{NLL-old} & \textbf{NLL-new} & \textbf{CRPS} & \textbf{NLL-old} & \textbf{NLL-new} \\
\midrule
\addlinespace[2pt]
\multicolumn{7}{l}{\emph{Standard Chaotic Dynamics}} \\
Rössler-Base         & \textbf{0.012}  & 0.016           & 0.024           & 0.039           & \textbf{0.010}  & 0.016           \\
Rössler-Param        & \textbf{0.040}  & 0.218           & 0.059           & \textbf{0.068}  & 3.16            & 0.086           \\
Lorenz-Base          & \textbf{0.759}  & 4.29            & 1.34            & \textbf{6.41}   & 14.8            & 11.6            \\
Lorenz-State         & \textbf{1.03}   & 3.24            & 1.82            & \textbf{4.14}   & 13.7            & 10.6            \\
Lorenz-Param         & 4.70            & 8.81            & \textbf{4.48}   & \textbf{18.2}   & 23.6            & 21.6            \\
Lorenz-Switch        & \textbf{2.27}   & 4.27            & 3.77            & \textbf{6.65}   & 15.1            & 11.8            \\
Lorenz96-Base        & -               & 0.635           & \textbf{0.436}  & -               & 2.50            & \textbf{2.29}   \\
Lorenz96-Switch      & -               & 2.44            & \textbf{1.58}   & -               & 7.29            & \textbf{7.06}   \\
Chua-Base            & \textbf{0.001}  & 0.001           & 0.006           & \textbf{0.011}  & 0.024           & 0.012           \\
Chua-Param           & 0.002           & \textbf{0.002}  & 0.008           & \textbf{0.007}  & 0.068           & 0.008           \\
Chua-Switch          & 0.003           & \textbf{0.002}  & 0.004           & \textbf{0.018}  & 0.041           & 0.019           \\
\addlinespace[2pt]
\multicolumn{7}{l}{\emph{Switching Linear Dynamical System}} \\
SLDS-Base            & -               & 2.73            & \textbf{2.55}   & -               & 2.97            & \textbf{2.89}   \\
SLDS-Param           & -               & 2.31            & \textbf{2.18}   & -               & 2.32            & \textbf{2.23}   \\
SLDS-Switch          & -               & 4.41            & \textbf{3.87}   & -               & 5.03            & \textbf{4.67}   \\
\addlinespace[2pt]
\multicolumn{7}{l}{\emph{Seasonal AR Shocks}} \\
SAR-Base             & \textbf{0.054}  & 0.067           & 0.063           & \textbf{0.053}  & 0.054           & 0.054           \\
SAR-Param            & \textbf{0.332}  & 0.365           & 0.342           & 0.335           & 0.338           & \textbf{0.326}  \\
\addlinespace[2pt]
\multicolumn{7}{l}{\emph{Double-Well Potential}} \\
DW-Base              & \textbf{0.048}  & 0.055           & 0.056           & \textbf{0.048}  & 0.068           & 0.063           \\
DW-Param             & 0.598           & \textbf{0.556}  & 0.624           & 0.754           & \textbf{0.716}  & 0.769           \\
DW-Switch            & 0.366           & 0.390           & \textbf{0.358}  & 0.675           & \textbf{0.634}  & 0.652           \\
\addlinespace[2pt]
\multicolumn{7}{l}{\emph{Ornstein--Uhlenbeck Diffusions}} \\
OU-Base              & -               & 0.197           & \textbf{0.195}  & -               & 0.207           & \textbf{0.203}  \\
OU-Param             & -               & \textbf{0.203}  & 0.217           & -               & 0.211           & \textbf{0.204}  \\
\bottomrule
\end{tabular}
\end{table*}

\begin{table*}[t]
\centering
\footnotesize
\caption{\textbf{Ablation Study: Forecasting CRPS (H=64/192).} Comparison of Fern variants across objective functions: CRPS loss (\texttt{CRPS}) with train sample = 8, evaluation samples = 96, the \textit{diagonal} Negative Log-Likelihood implementation (\texttt{NLL-old}) with train sample = 8, evaluation samples = 96, and Fern Gaussian NLL with full covariance matrix (\texttt{NLL-new}) with train sample = 4, evaluation samples = 32. As expected, directly minimizing CRPS often yields the tightest distributional scores, though the multivariate Gaussian NLL formulations strictly enforce geometric covariance structure.}
\label{app:tab:ablation-crps}
\setlength{\tabcolsep}{4.0pt}
\begin{tabular}{l ccc ccc}
\toprule
 & \multicolumn{3}{c}{\textbf{H=64}} & \multicolumn{3}{c}{\textbf{H=192}} \\
\cmidrule(lr){2-4} \cmidrule(lr){5-7}
\textbf{Dataset} & \textbf{CRPS} & \textbf{NLL-old} & \textbf{NLL-new} & \textbf{CRPS} & \textbf{NLL-old} & \textbf{NLL-new} \\
\midrule
\addlinespace[2pt]
\multicolumn{7}{l}{\emph{Standard Chaotic Dynamics}} \\
Rössler-Base         & \textbf{0.054}  & 0.062           & 0.090           & 0.082           & \textbf{0.046}  & 0.071           \\
Rössler-Param        & \textbf{0.070}  & 0.089           & 0.095           & \textbf{0.083}  & 0.293           & 0.144           \\
Lorenz-Base          & \textbf{0.351}  & 0.759           & 0.555           & \textbf{0.814}  & 1.32            & 1.47            \\
Lorenz-State         & \textbf{0.345}  & 0.586           & 0.621           & \textbf{0.641}  & 1.23            & 1.54            \\
Lorenz-Param         & \textbf{0.674}  & 1.00            & 0.707           & \textbf{1.49}   & 1.83            & 2.23            \\
Lorenz-Switch        & \textbf{0.389}  & 0.638           & 0.908           & \textbf{0.763}  & 1.32            & 1.57            \\
Lorenz96-Base        & -               & 0.305           & \textbf{0.295}  & -               & \textbf{0.723}  & 0.823           \\
Lorenz96-Switch      & -               & 0.667           & \textbf{0.611}  & -               & \textbf{1.36}   & 1.65            \\
Chua-Base            & \textbf{0.024}  & 0.025           & 0.046           & \textbf{0.044}  & 0.046           & 0.053           \\
Chua-Param           & 0.030           & \textbf{0.025}  & 0.050           & \textbf{0.044}  & 0.058           & 0.052           \\
Chua-Switch          & 0.030           & \textbf{0.029}  & 0.042           & \textbf{0.051}  & 0.062           & 0.066           \\
\addlinespace[2pt]
\multicolumn{7}{l}{\emph{Switching Linear Dynamical System}} \\
SLDS-Base            & -               & 1.02            & \textbf{1.00}   & -               & \textbf{1.04}   & 1.06            \\
SLDS-Param           & -               & \textbf{0.906}  & 0.962           & -               & \textbf{0.914}  & 0.974           \\
SLDS-Switch          & -               & 1.24            & \textbf{1.20}   & -               & 1.39            & \textbf{1.31}   \\
\addlinespace[2pt]
\multicolumn{7}{l}{\emph{Seasonal AR Shocks}} \\
SAR-Base             & \textbf{0.134}  & 0.148           & 0.144           & \textbf{0.132}  & 0.134           & 0.134           \\
SAR-Param            & \textbf{0.332}  & 0.350           & 0.343           & \textbf{0.333}  & 0.335           & 0.336           \\
\addlinespace[2pt]
\multicolumn{7}{l}{\emph{Double-Well Potential}} \\
DW-Base              & \textbf{0.124}  & 0.142           & 0.135           & \textbf{0.123}  & 0.169           & 0.144           \\
DW-Param             & 0.436           & \textbf{0.403}  & 0.469           & \textbf{0.487}  & 0.490           & 0.575           \\
DW-Switch            & \textbf{0.305}  & 0.331           & 0.322           & 0.449           & \textbf{0.448}  & 0.504           \\
\addlinespace[2pt]
\multicolumn{7}{l}{\emph{Ornstein--Uhlenbeck Diffusions}} \\
OU-Base              & -               & \textbf{0.256}  & 0.263           & -               & \textbf{0.267}  & 0.268           \\
OU-Param             & -               & \textbf{0.259}  & 0.279           & -               & \textbf{0.264}  & 0.268           \\
\bottomrule
\end{tabular}
\end{table*}
 
\section*{App: Statistical Properties of the Whitened Mahalanobis Diagnostic}
\label{app:mahalanobis}

When evaluating the calibration of probabilistic forecasters, the Mahalanobis distance of the residuals serves as a standard diagnostic. However, calculating this distance using individual model samples versus the ensemble mean yields fundamentally different statistical expectations, particularly in the presence of post-hoc noise injection.

Let $y_{\text{clean}}$ denote the deterministic dynamics of the underlying physical system. For noisy environments (e.g., measurement error), the observed target is $y = y_{\text{clean}} + \epsilon$, where the aleatoric noise is distributed as $\epsilon \sim \mathcal{N}(0, \Sigma_{\text{true}})$. 
Our generative model predicts a distribution from which we draw samples $\hat{y} = \mu + z$, where $\mu$ is the model's predictive mean and the sampling noise is $z \sim \mathcal{N}(0, \Sigma_{\text{model}})$. We assume a perfectly calibrated model such that $\mu = y_{\text{clean}}$ and $\Sigma_{\text{model}} = \Sigma_{\text{true}} = \Sigma$.

\subsection*{1 The Sample-Based Variance Inflation}
If we compute the residual using an individual model sample, $r = y - \hat{y}$, we obtain:
\begin{equation}
    r = (y_{\text{clean}} + \epsilon) - (\mu + z) = \epsilon - z
\end{equation}
Because the physical measurement noise $\epsilon$ and the model's internal sampling noise $z$ are strictly independent, their variances sum linearly. The true distribution of the sample-based residual is therefore $r \sim \mathcal{N}(0, 2\Sigma)$. 

We evaluate calibration by computing the expected squared Mahalanobis distance, $D_M^2$, over the spatial and feature dimensions $P$. By the linearity of expectation and the cyclic property of the trace:
\begin{equation}
    \mathbb{E}[D_M^2] = \mathbb{E}[r^T \Sigma^{-1} r] = \mathbb{E}[\epsilon^T \Sigma^{-1} \epsilon] + \mathbb{E}[z^T \Sigma^{-1} z]
\end{equation}
The cross-term $\mathbb{E}[\epsilon^T \Sigma^{-1} z]$ vanishes due to independence. Expanding the expectations yields $\text{tr}(\Sigma^{-1}\Sigma) + \text{tr}(\Sigma^{-1}\Sigma) = P + P = 2P$. Thus, for noisy targets, the expected distance is artificially inflated by a factor of 2.

\subsection*{2 The Ensemble Mean Correction}
To resolve this duality and align with the formal definition of the Mahalanobis metric—which measures the distance from an observation to the center of a distribution—we define the residual using the ensemble mean: $r_{\text{mean}} = y - \mu$. 

Under this formulation, the model's sampling noise $z$ is marginalized out. For noisy targets, the residual strictly isolates the physical noise: $r_{\text{mean}} = \epsilon$, yielding $\mathbb{E}[D_M^2] = P$. This formally isolates the true aleatoric noise geometry, allowing the whitened residuals $\Sigma^{-1/2}(y - \mu)$ to be rigorously tested against the standard normal $\mathcal{N}(0, I)$ without variance inflation.

\section{App: Configuration of shock experiments }
We use \emph{exactly the same} shock setup as in \citet{wang2025friren} for direct comparability. We record the state after every integration step, so the sampling interval of the time series equals the solver step size $dt$. Since the train/validation/test split is $0.7/0.2/0.1$, setting \texttt{shock\_frac}$=0.35$ (fraction of the full trajectory) places the shock at $50\%$ of the training segment ($0.35/0.7=0.5$).

\begin{table*}[t]
\centering
\scriptsize
\caption{Parameter settings for the main chaotic benchmarks (top block) and all synthetic shock scenarios used in our experiments (bottom block). Shock scenarios are instantiated via \texttt{PremadeID.*} in \texttt{make\_source}, with \texttt{shock\_frac} fixed to $0.35$ so that shocks are applied after the first 35\% of the trajectory.}
\label{tab:app:shock-params}
\setlength{\tabcolsep}{4pt}
\renewcommand{\arraystretch}{1.1}
\begin{tabular}{llclp{8.3cm}}
\toprule
System & Scenario & $dt$ & steps & Parameters / shock description \\
\midrule
\multicolumn{5}{l}{\textit{Main chaotic benchmarks (no shock).}}\\
Lorenz-63 & main & $0.01$ & $25000$ &
$\texttt{sigma}=10,\ \texttt{rho}=28,\ \texttt{beta}=8/3$. \\[2pt]
R\"ossler & main & $0.01$ & $25000$ &
$\texttt{a}=0.2,\ \texttt{b}=0.2,\ \texttt{c}=5.7$. \\[2pt]
Chua & main & $0.005$ & $35000$ &
$\texttt{alpha}=15.6,\ \texttt{beta}=28.0,\ \texttt{m0}=-8/7,\ \texttt{m1}=-5/7$. \\
\midrule
\multicolumn{5}{l}{\textit{Synthetic shock scenarios (code identifiers \texttt{PremadeID.*}).}}\\
Lorenz-63 & \texttt{LORENZ\_BASE} & $0.01$ & $35999$ &
Baseline Lorenz-63, no shock; default $\texttt{sigma}=10,\ \texttt{rho}=28,\ \texttt{beta}=8/3$. \\[2pt]
Lorenz-63 & \texttt{LORENZ\_PARAM} & $0.01$ & $35999$ &
Parameter shock (\texttt{shock\_kind = "param"}):
$\texttt{sigma}: 10 \to 10.1,\ \texttt{rho}: 28 \to 28.1,\ \texttt{beta}: 8/3 \to 8.1/3$. \\[2pt]
Lorenz-63 & \texttt{LORENZ\_STATE} & $0.01$ & $35999$ &
State shock (\texttt{shock\_kind = "state\_eps"}):
$\texttt{shock\_eps} = 0.9$; ODE parameters as in \texttt{LORENZ\_BASE}. \\[2pt]
Lorenz-63 & \texttt{LORENZ\_SWITCH} & $0.01$ & $35999$ &
Switch shock (\texttt{shock\_kind = "switch"}):
\texttt{switch\_update} sets $\texttt{rho}: 28 \to 28.1$ and
$\texttt{initial\_cond}: [1.0, 0.98, 1.1] \to [1.002, 0.982, 1.102]$. \\[2pt]

R\"ossler & \texttt{ROSSLER\_BASE} & $0.01$ & $35999$ &
Baseline R\"ossler, no shock; $\texttt{a}=0.2,\ \texttt{b}=0.2,\ \texttt{c}=5.7$. \\[2pt]
R\"ossler & \texttt{ROSSLER\_PARAM} & $0.01$ & $35999$ &
Parameter shock (\texttt{shock\_kind = "param"}):
$\texttt{a}: 0.2 \to 0.25,\ \texttt{b}: 0.2 \to 0.25,\ \texttt{c}: 5.7 \to 5.75$. \\[2pt]

Lorenz-96 & \texttt{LORENZ96\_BASE} & $0.007$ & $55000$ &
Baseline Lorenz-96; $\texttt{dim} = 6$, $\texttt{forcing} = 8.0$,
$\texttt{method} = \text{"rk4"}$. \\[2pt]
Lorenz-96 & \texttt{LORENZ96\_SWITCH} & $0.007$ & $55000$ &
Switch shock (\texttt{shock\_kind = "switch"}):
\texttt{switch\_update} sets $\texttt{forcing}: 8.0 \to 9.0$ and
$\texttt{initial\_cond} = [0.99, 1.02, 1.02, 1.03, 1.01, 1.01]$
(with $\texttt{dim} = 6$, \texttt{method}=\text{"rk4"}). \\[2pt]

Chua & \texttt{CHUA\_BASE} & $0.005$ & $35999$ &
Baseline Chua, no shock; $\texttt{alpha}=15.6,\ \texttt{beta}=28.0,\ \texttt{m0}=-8/7,\ \texttt{m1}=-5/7$. \\[2pt]
Chua & \texttt{CHUA\_PARAM} & $0.005$ & $35999$ &
Parameter shock (\texttt{shock\_kind = "param"}):
$\texttt{alpha}: 15.6 \to 15.9,\ \texttt{beta}: 28.0 \to 28.5,\ 
\texttt{m0}: -8/7 \to -8.1/7,\ \texttt{m1}: -5/7 \to -5.2/7$. \\[2pt]
Chua & \texttt{CHUA\_SWITCH} & $0.005$ & $35999$ &
Switch shock (\texttt{shock\_kind = "switch"}):
\texttt{switch\_update} sets
$\texttt{initial\_cond} = [0.11, 0.01, 0.02]$; other parameters as in \texttt{CHUA\_BASE}. \\[2pt]

OU & \texttt{OU\_BASE} & $0.5$ & $25000$ &
Baseline Ornstein--Uhlenbeck;
$\texttt{initial\_cond}=[0.0],\ \texttt{theta}=0.2,\ \texttt{mu}=0.0,\ \texttt{sigma}=0.3,\ \texttt{method}$ = \text{"euler"}. \\[2pt]
OU & \texttt{OU\_PARAM} & $0.5$ & $25000$ &
Parameter shock (\texttt{shock\_kind = "param"}):
$\texttt{mu}: 0.0 \to 0.5$; other OU parameters as in \texttt{OU\_BASE}. \\[2pt]

SLDS & \texttt{SLDS\_BASE} & $0.01$ & $25000$ &
Baseline switching linear dynamical system;
$\texttt{A1}=0.9,\ \texttt{Q1}=0.05,\ \texttt{A2}=0.98,\ \texttt{Q2}=0.35,\ \texttt{p11}=0.94,\ \texttt{p22}=0.95$. \\[2pt]
SLDS & \texttt{SLDS\_PARAM} & $0.01$ & $25000$ &
Parameter shock (\texttt{shock\_kind = "param"}):
$\texttt{A1}: 0.9 \to 0.83,\ \texttt{Q1}: 0.05 \to 0.50,\ 
\texttt{A2}: 0.98 \to 0.97,\ \texttt{Q2}: 0.35 \to 0.30,\ 
\texttt{p11}: 0.94 \to 0.96,\ \texttt{p22}: 0.95 \to 0.92$. \\[2pt]
SLDS & \texttt{SLDS\_SWITCH} & $0.01$ & $25000$ &
Switch shock (\texttt{shock\_kind = "switch"}):
\texttt{switch\_update} sets
$\texttt{A1}=0.87,\ \texttt{Q1}=0.07,\ \texttt{A2}=0.99,\ \texttt{Q2}=0.45,\ \texttt{p11}=0.90,\ \texttt{p22}=0.95$. \\[2pt]

Double-well & \texttt{DOUBLEWELL\_BASE} & $0.5$ & $25000$ &
Baseline double-well SDE (Euler API, step size $0.5$);
$\texttt{a}=1.5,\ \texttt{sigma}=0.25,\ \texttt{seed}=1955$. \\[2pt]
Double-well & \texttt{DOUBLEWELL\_PARAM} & $0.5$ & $25000$ &
Parameter shock (\texttt{shock\_kind = "param"}):
$\texttt{a}: 1.5 \to 1.0,\ \texttt{sigma}: 0.25 \to 0.35$. \\[2pt]
Double-well & \texttt{DOUBLEWELL\_SWITCH} & $0.5$ & $25000$ &
Switch shock (\texttt{shock\_kind = "switch"}):
\texttt{switch\_update} sets $\texttt{a}=1.0,\ \texttt{sigma}=0.35$
(same target as \texttt{DOUBLEWELL\_PARAM}). \\[2pt]

Seasonal AR & \texttt{SEASONAL\_AR\_BASE} & $0.01$ & $25000$ &
Baseline seasonal AR process (discrete-time; \texttt{dt}/\texttt{method} kept for API):
$\texttt{S}=24,\ \texttt{phi}=0.5,\ \texttt{sigma}=0.2,\ \texttt{a0}=1.0,\ \texttt{amp\_drift\_per\_step}=0$. \\[2pt]
Seasonal AR & \texttt{SEASONAL\_AR\_PARAM} & $0.01$ & $25000$ &
Parameter shock (\texttt{shock\_kind = "param"}):
$\texttt{a0}: 1.0 \to 1.4,\ \texttt{sigma}: 0.2 \to 0.35,\ \texttt{phi}: 0.5 \to 0.8$. \\[2pt]

GARCH(1,1) & \texttt{GARCH\_BASE} & $0.01$ & $25000$ &
Baseline GARCH(1,1) volatility model (discrete-time; \texttt{dt}/\texttt{method} kept for API):
$\texttt{omega}=0.01,\ \texttt{alpha}=0.06,\ \texttt{beta}=0.90$. \\[2pt]
GARCH(1,1) & \texttt{GARCH\_PARAM} & $0.01$ & $25000$ &
Parameter shock (\texttt{shock\_kind = "param"}):
$\texttt{omega}: 0.01 \to 0.03,\ \texttt{alpha}: 0.06 \to 0.15,\ \texttt{beta}: 0.90 \to 0.70$. \\[2pt]

KS & \texttt{KS\_BASE} & $0.01$ & $25000$ &
Baseline Kuramoto--Sivashinsky;
$\texttt{nx}=64,\ \texttt{Lx}=22.0,\ \texttt{nu}=1.0,\ \texttt{method}$ = \text{"etdrk4"}. \\[2pt]
KS & \texttt{KS\_PARAM} & $0.01$ & $25000$ &
Parameter shock (\texttt{shock\_kind = "param"}):
$\texttt{nu}: 1.0 \to 0.80$ (other KS parameters as in \texttt{KS\_BASE}). \\
\bottomrule
\end{tabular}
\end{table*}

\section{App: Systems and Datasets}
\label{app:data}
Here we provide a short introduction to the systems.  

\subsection{Chaotic systems}
Chaotic dynamics are sensitive to discretization and numerical precision: finite precision can suppress chaos and induce spurious periodic orbits, and truncation/round-off errors can dominate long-horizon trajectories. We therefore follow the exact data-generation protocol of \citet{wang2025friren}: all chaotic ODE datasets (Lorenz-63, R\"ossler, Chua, Lorenz-96) are simulated in \texttt{float64} using a 4th-order Runge--Kutta (RK4) integrator, and only then converted to \texttt{float32} PyTorch tensors for training and evaluation. 

\paragraph{Lorenz-63.}
A canonical three-dimensional model of atmospheric convection:
\[
  \dot x = \sigma(y-x),\qquad
  \dot y = x(\rho-z)-y,\qquad
  \dot z = xy-\beta z.
\]

\paragraph{R\"ossler.}
A three-dimensional system exhibiting a folded-band attractor:
\[
  \dot x = -y-z,\qquad
  \dot y = x + a y,\qquad
  \dot z = b + z(x-c).
\]

\paragraph{Chua's circuit.}
A three-dimensional piecewise-linear circuit model with a double-scroll attractor:
\[
  \dot x = \alpha\bigl(y-x-h(x)\bigr),\qquad
  \dot y = x-y+z,\qquad
  \dot z = -\beta y,
\]
where
\[
  h(x) = m_1 x + \tfrac{1}{2}(m_0-m_1)\bigl(|x+1|-|x-1|\bigr).
\]

\paragraph{Lorenz-96.}
A $d$-dimensional toy model for mid-latitude atmospheric dynamics with cyclic nearest-neighbour coupling and constant forcing $F$ (\texttt{forcing} in code, dimension \texttt{dim}):
\[
  \dot x_j = (x_{j+1}-x_{j-2})\,x_{j-1} - x_j + F,\qquad j=1,\dots,d,
\]
with indices taken modulo $d$ (e.g.\ $x_0=x_d$ and $x_{-1}=x_{d-1}$).

\subsection{Stochastic systems}
\paragraph{Ornstein--Uhlenbeck (OU).}
A one-dimensional mean-reverting diffusion with parameters $(\theta,\mu,\sigma)$ (\texttt{theta}, \texttt{mu}, \texttt{sigma} in code):
\[
  dX_t = \theta(\mu-X_t)\,dt + \sigma\,dW_t,
\]
simulated with an Euler--Maruyama scheme at step size \texttt{dt}.

\paragraph{Double-well SDE.}
A one-dimensional bistable diffusion in a double-well potential, parameterized by a shape parameter $a$ and noise scale $\sigma$ (\texttt{a}, \texttt{sigma} in code):
\[
  dX_t = (aX_t - X_t^3)\,dt + \sigma\,dW_t.
\]
The drift $aX_t-X_t^3$ induces two metastable wells around $\pm\sqrt{a}$, with noise-driven transitions between wells.

\paragraph{Switching linear (SLDS).}
A one-dimensional switching linear dynamical system with two linear-Gaussian regimes $(A_1,Q_1)$ and $(A_2,Q_2)$ and Markov self-transition probabilities $p_{11},p_{22}$ (\texttt{A1}, \texttt{Q1}, \texttt{A2}, \texttt{Q2}, \texttt{p11}, \texttt{p22} in code). Let $s_t\in\{1,2\}$ denote the latent regime:
\[
  x_{t+1} = A_{s_t}x_t + \eta_t,\qquad \eta_t\sim\mathcal N(0,Q_{s_t}),
\]
\[
  \mathbb P[s_{t+1}=i\mid s_t=i]=p_{ii},\quad i\in\{1,2\}.
\]
This yields piecewise-linear dynamics with switches driven by a two-state Markov chain.

\paragraph{Seasonal AR.}
A one-dimensional discrete-time process combining an AR(1) term and a seasonal component of period $S$ (\texttt{S}) with slowly drifting amplitude. With AR coefficient $\phi$ (\texttt{phi}), innovation scale $\sigma$ (\texttt{sigma}), initial amplitude $a_0$ (\texttt{a0}), and linear drift rate \texttt{amp\_drift\_per\_step}:
\[
  a_t = a_0 + t\cdot \mathrm{amp\_drift\_per\_step},
\]
\[
  x_t = a_t \cos\!\Bigl(\tfrac{2\pi t}{S}\Bigr) + \phi x_{t-1} + \sigma\,\varepsilon_t,\qquad
  \varepsilon_t\sim\mathcal N(0,1).
\]

\paragraph{GARCH(1,1).}
A discrete-time volatility model with conditionally Gaussian returns and autoregressive conditional variance, parameterized by $(\omega,\alpha,\beta)$ (\texttt{omega}, \texttt{alpha}, \texttt{beta} in code):
\[
  x_t = \sigma_t \varepsilon_t,\qquad \varepsilon_t\sim\mathcal N(0,1),
\]
\[
  \sigma_t^2 = \omega + \alpha x_{t-1}^2 + \beta \sigma_{t-1}^2.
\]
Here $x_t$ represents volatility-clustered returns, while $\sigma_t^2$ evolves as a GARCH(1,1) variance process.

\section{APP: Fern Settings} 
\label{app:lit:arch} 

\paragraph{Overall structure.}
Fern consists of (i) a bidirectional encoder that couples the context embedding with a latent Gaussian $z$ through $5$ coupling rounds, (ii) a short processor that couples $z$ with prediction-space noise, and (iii) a patchwise Brenier decoder that generates the affine map parameters $(t_y,U,\Lambda)$ and applies the shift--rotate--scale--rotate-back transport in parallel over horizon patches.

\paragraph{Soft-bounded coefficients.}
For numerical stability, raw network outputs that parameterize translations and scales are passed through differentiable soft bounds: the mapping is approximately linear inside a target interval $[\ell,u]$ and saturates smoothly outside, preventing exploding updates while retaining gradient flow near the boundaries.

\paragraph{SPD scaling and translation bounds.}
The Brenier scale acts multiplicatively as $(1+c)\odot y$, so bounding $c\in[-1,4.5]$ yields effective eigenvalues in $[0,5.5]$ and avoids degenerate ellipsoids. Translations use wider bounds (e.g., $t_y\in[-15,15]$) to accommodate large mean shifts without destabilizing training.

\paragraph{Complex (block-diagonal) scaling in the encoder.}
To capture low-cost rotational behavior in the latent-to-context couplings, some encoder layers use a $2\times2$ block-diagonal ``complex'' scaling (interleaving real/imaginary pairs), while others use diagonal scaling. This alternation balances expressivity and gradient stability in affine coupling updates.

\paragraph{Householder rotations.}
Orthogonal rotations $U$ are parameterized as a product of $R$ Householder reflections per patch. Unless otherwise stated we use $R=24$ (full-capacity rotations). For memory efficiency, we optionally train only a contiguous block of reflections per step via stop-gradient scheduling, while cycling blocks across iterations; this reduces backpropagation cost without changing the forward map.

\bibliography{uai2026-template}

\end{document}